\documentclass[sigconf, screen, preprint]{acmart}

\usepackage{tcolorbox}
\usepackage{graphicx}
\usepackage{caption}
\usepackage{subcaption}
\usepackage{tabularx}
\usepackage{url}
\usepackage{booktabs}
\usepackage{multicol}
\usepackage{multirow}
\usepackage{enumitem}
\usepackage{algorithm}
\usepackage{algpseudocode}
\usepackage{soul}
\AtBeginDocument{%
  }

\setcopyright{none}              
\renewcommand\footnotetextcopyrightpermission[1]{} 

\acmConference{}{}{}
\acmBooktitle{}
\acmPrice{}
\acmDOI{}
\acmISBN{}
\copyrightyear{}
\acmYear{}

\title{\ SteerDiff: Steering towards Safe Text-to-Image Diffusion Models}

\author{Hongxiang Zhang}
\email{hxxzhang@ucdavis.edu}
\affiliation{%
  \institution{University of California, Davis}
  \city{Davis}
  \state{California}
  \country{USA}
}

\author{Yifeng He}
\email{yfhe@ucdavis.edu}
\affiliation{%
  \institution{University of California, Davis}
  \city{Davis}
  \state{California}
  \country{USA}
\email{yfeng@ucdavis.edu}
}

\author{Hao Chen}
\email{chen@ucdavis.edu}
\affiliation{%
  \institution{University of California, Davis}
  \city{Davis}
  \state{California}
  \country{USA}
\email{chen@ucdavis.edu}
}

\renewcommand{\shortauthors}{Hongxiang et al.}
\newcommand\safediff{SteerDiff}

\begin{document}

\begin{abstract}
Text-to-image (T2I) diffusion models have drawn attention for their ability to generate high-quality images with precise text alignment. However, these models can also be misused to produce harmful or inappropriate content. Existing safety measures, which typically rely on text classifiers or ControlNet-like approaches, are often insufficient.
Text classifiers rely on large-scale labeled datasets and can be easily bypassed by rephrasing.
As diffusion models continue to scale, fine-tuning safeguards becomes increasingly challenging and lacks flexibility. Recent red-teaming attack researches further underscore the need for a new paradigm to prevent the generation of inappropriate content.
In this paper, we introduce {\safediff}, a lightweight adapter module designed to act as an intermediary between user input and the diffusion model, ensuring that generated images adhere to ethical and safety standards with little to no impact on usability. {\safediff} identifies and manipulates inappropriate concepts within the text embedding space to guide the model away from harmful outputs. We conducted extensive experiments across multiple concept unlearning tasks to evaluate the effectiveness of our approach. Furthermore, we benchmark {\safediff} against multiple red-teaming strategies to assess its robustness. Finally, we explore the potential of {\safediff} for artist-style removal tasks, demonstrating its versatility in text-conditioned image generation.

\textcolor{red}{\textbf{Warning: This paper contains potentially offensive text and images.}}
\end{abstract}





\maketitle

\section{Introduction}
\label{sec:intro}

Text-to-image~(T2I) diffusion models have attracted attention for their out-of-the-box functionality and the superior quality of their generated images. Using models like Stable Diffusion~\citep{rombach2022high,rombach2021highresolution} or DALL-E~\citep{ramesh2022hierarchical}, users can use simple natural language descriptions as input to generate high-quality images with precise text alignment. This capability is largely contributed by pre-trained language models~\citep{hammoud2024synthclip,jin2020bert,dosovitskiy2020image} that learn and reflect the underlying syntax and semantics, as well as by extensive multimodal training datasets that encompass a wide range of text-to-image aligned content. However, these training methods also introduce the risk of generating inappropriate content with unsafe concepts such as sex, violence, or illegal activity. Thus, preventing the generation of inappropriate images is both critical and urgent.

To tackle these challenges, recent research has integrated several safety strategies to prevent inappropriate content generation. 
Model editing methods~\citep{gandikota2023erasing,orgad2023editing,gandikota2024unified} identify and prune weights within the diffusion model to remove harmful concepts, but they often degrade output quality, thus harming the utility of the models~\citep{rombach2022high}.
Unlearning methods\citep{zhang2024forget,huang2024receler} finetune models to erase harmful content, but as model sizes grow, this becomes increasingly computationally expensive and less practical\citep{ramesh2022hierarchical,rombach2022high,saharia2022photorealistic}.
A practical alternative lies in training-free, filtering-based methods, which aim to exclude unsafe concepts directly from input prompts without modifying the underlying model. However, existing approaches in this category~\citep{liu2024latent,schuhmann2022laion,schramowski2023safe,rando2022red} can be easily passed by text-based attacks~\citep{li2018textbugger,jin2020bert,garg2020bae,maus2023black} through mislead the classification mechanisms in the filter by rephrasing, e.g.``\textit{A photo of a billboard showing a naked man.}'' into ``\textit{A photo of a billboard showing an LGBT man in an explicit position}''.


Although the aforementioned safety mechanisms have shown effectiveness according to their respective evaluation schemes, recent red-teaming studies demonstrate their potential flaws~\citep{zhang2023generate,qu2023unsafe}. \citet{chin2023prompting4debugging} show that approximately half of the prompts, which were originally blocked by existing safety mechanisms, can be manipulated by their Prompting4Debugging~(P4D) to become problematic. Similarly, Unsafe Diffusion~\citep{qu2023unsafe} finds that 14.56\% of generated images across four state-of-the-art T2I models and their prompt datasets were unsafe, underscoring the vulnerability of these models to generating harmful content.
Moreover, black-box jailbreaking approaches Jailbreak Prompt Attack (JPA) and SneakyPrompt~\citep{ma2024jailbreaking,yang2024sneakyprompt} successfully attack both online services and offline T2I models with the current safety mechanisms. 

\begin{figure*}[ht]
    \centering
    \includegraphics[width=\textwidth]{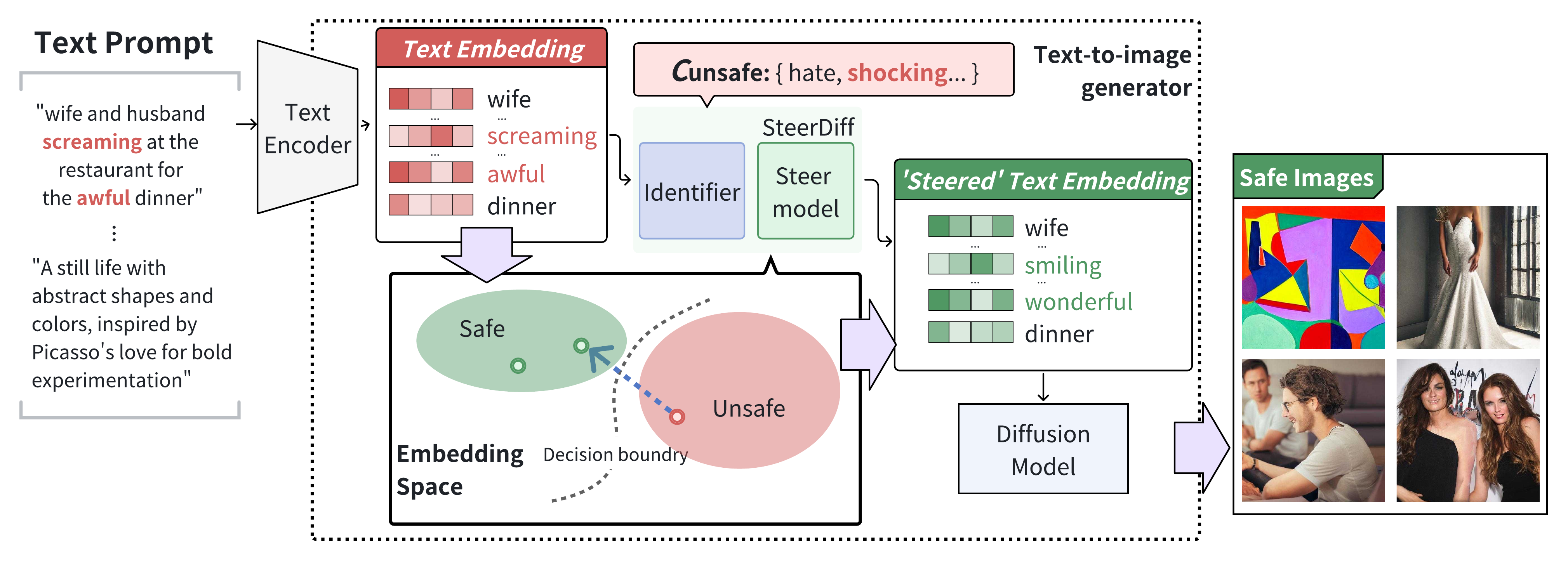}
    \caption{Overview of {\safediff}, a safety method designed to identify and steer inappropriate concepts toward producing safe images. The input prompt is first embedded by a text encoder. The identifier then checks if the prompt contains any inappropriate concepts $c' \in C_{\text{unsafe}}$. If detected, a linear transformation is applied to the prompt’s embedding to steer it toward safer content. This transformation adjusts the embedding space while preserving the semantics of the original prompt. Once transformed, the modified embedding is passed through the diffusion model to generate safe images.}
    \label{F:overviewsafe}
\end{figure*}

In this work, we propose {\safediff}, a novel lightweight adaptor model for text-conditioned diffusion models to ensure safe generation without altering model weights. As illustrated in \autoref{F:overviewsafe}, {\safediff} first constructs a semantic boundary distinguishing between safe and unsafe content. Then, it projects potentially unsafe embeddings toward the safe region while preserving the original semantics and maintaining the diffusion model's generative capabilities. This approach offers three key advantages: efficiency, effectiveness, and versatility. By operating at the prompt embedding level, our method eliminates the need for computationally intensive model re-training while preserving the original semantics. Moreover, by filtering unsafe content early in the generation process, {\safediff} prevents harmful latent representations and achieves state-of-the-art robustness in our red-teaming experiments.

We benchmark {\safediff} against state-of-the-art concept removal techniques, including Erased Stable Diffusion~(ESD)~\citep{gandikota2023erasing}, SPM~\cite{lyu2024one} and Safe Latent Diffusion (SLD)~\citep{schramowski2023safe,rando2022red}, for removing inappropriate content. Experimental results demonstrate that {\safediff} significantly reduces inappropriate content generation while preserving image quality and semantic fidelity. Furthermore, we evaluate its robustness by defending against red-teaming methods, including Prompt4Debugging, Ring-A-Bell, MMA-Diffusion, and SneakyPrompt\cite{chin2023prompting4debugging,tsai2023ring,yang2024mma,yang2024sneakyprompt}, demonstrating the effectiveness of our approach in mitigating various forms of adversarial attacks on text-to-image generation systems. Lastly, we evaluate the potential of {\safediff} for artist-style removal tasks, highlighting its versatility in text-conditioned image generation. In summary, our main
contributions are as follows:
\begin{itemize}
    \item We propose a novel lightweight adaptor model {\safediff} for T2I diffusion models to ensure safe image generation without altering the model weights.
    \item We comprehensively evaluate {\safediff} across different scenarios and red-teaming attacks, achieving state-of-the-art robustness, semantic consistency, and image quality.
    \item {\safediff} shows versatility for other tasks, such as artist-style removal, underscoring its broad applicability in text-conditioned image generation scenarios.
\end{itemize}







\section{Methodology}
\label{sec:method}

Determining what constitutes inappropriate imagery is highly subjective and varies based on context, setting, cultural and social predispositions, and individual factors. Additionally, \cite{qu2023unsafe} observes that unsafe content can be shared through memes. In the context of T2I models, these new concepts are embedded within the words of a prompt. Consequently, adding new concepts to block T2I generation without retraining the diffusion model is impractical. 
To overcome these limitations, we formalize the problem by identifying unsafe concepts in text embedding space and then projecting potentially unsafe embeddings toward safe regions. This projection preserves the original semantics while maintaining the diffusion model's generative capabilities. Our approach allows us to define blacklisted concepts at test time, enabling greater flexibility.

In \autoref{traindatacollect}, we find a set of inappropriate concepts based on established work~\citep{schramowski2023safe} and describe the process of collecting and generating training data using a large language model as illustrated in \autoref{F:ar}. As shown in the overview in \autoref{F:overviewsafe}, we next explain how inappropriate concepts are identified within text embeddings (\autoref{identifier}), followed by our steering approach to mitigate the generation of inappropriate content (\autoref{steer}). Finally, we outline how the framework operates during inference to identify and steer text prompts associated with unsafe concepts (\autoref{inference}).

\vspace{-0.15cm}
\subsection{Training Data Collection}
\label{traindatacollect}
    Directly classifying safe/unsafe prompts requires large-scale annotated datasets~\citep{markov2023holistic}.
    Following Safe Latent Diffusion, we base our definition of inappropriate content on the work of \citeauthor{gebru2021datasheets}: ``[data that] \textit{if viewed directly, might be offensive, insulting, threatening, or might otherwise cause anxiety}''~\citep{gebru2021datasheets,schramowski2023safe}. 
    Specifically, we consider an image as inappropriate if it contains any concept $c \in C_{\text{unsafe}}$, where
    \begin{equation}
        \begin{aligned}
        C_{\text{unsafe}} = & \{\text{hate, harassment, violence, self-harm, sexual } \\
        & \text{content, shocking images, illegal activity}\}
        \label{unsafeconcept}
        \end{aligned}
    \end{equation}

    It is important to note that the definition of inappropriateness is not restricted to these seven categories, as the boundaries of appropriateness vary across cultures and evolve over time. In this study, we limit our scope to images displaying clear and tangible inappropriate behavior acts.
    Although we limit our scope to the current $C_{\text{unsafe}}$, our framework can be extended beyond this task, as discussed in \autoref{L:conceptforg}.

\begin{figure*}[ht]
        \centering
    \begin{subfigure}[t]{0.55\textwidth}
    \centering
    \includegraphics[width=\textwidth]{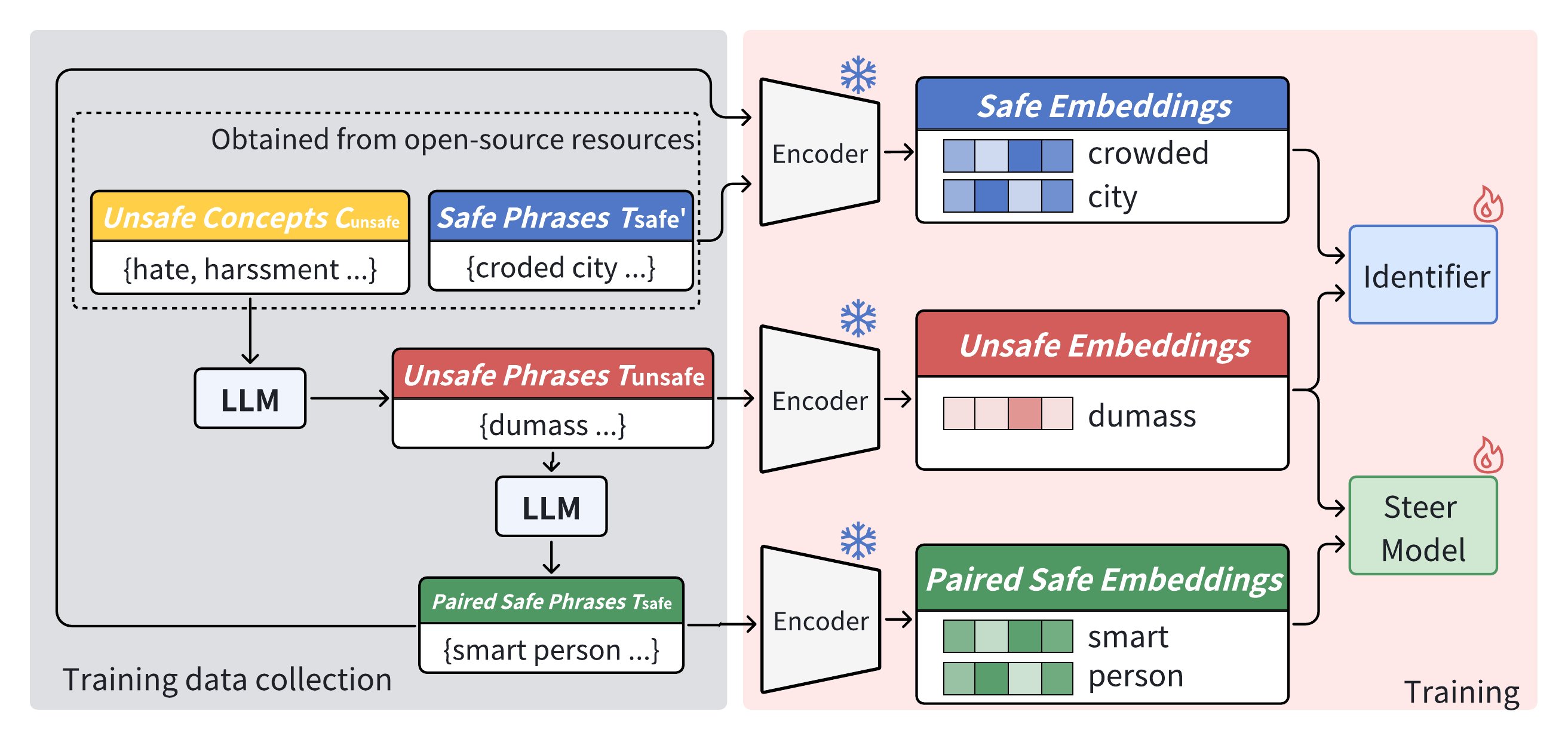}
    \caption{Overview of data collection and training process: We begin with a defined set of unsafe concepts, $C_{\text{unsafe}}$ (yellow block). Next, we use an LLM to generate related unsafe (red block on the left) and safe (green block on the left) phrases based on each concept $c \in C_{\text{unsafe}}$. Additionally, we incorporated a set of safe phrases from open sources~(blue block) to serve as a part of safe embeddings. These phrases are then encoded to train the identifier and steering model.}
    \label{F:ar} 
    \end{subfigure}
    \hfill
    \begin{subfigure}[t]{0.44\textwidth}
    \centering
    \includegraphics[width=\textwidth]{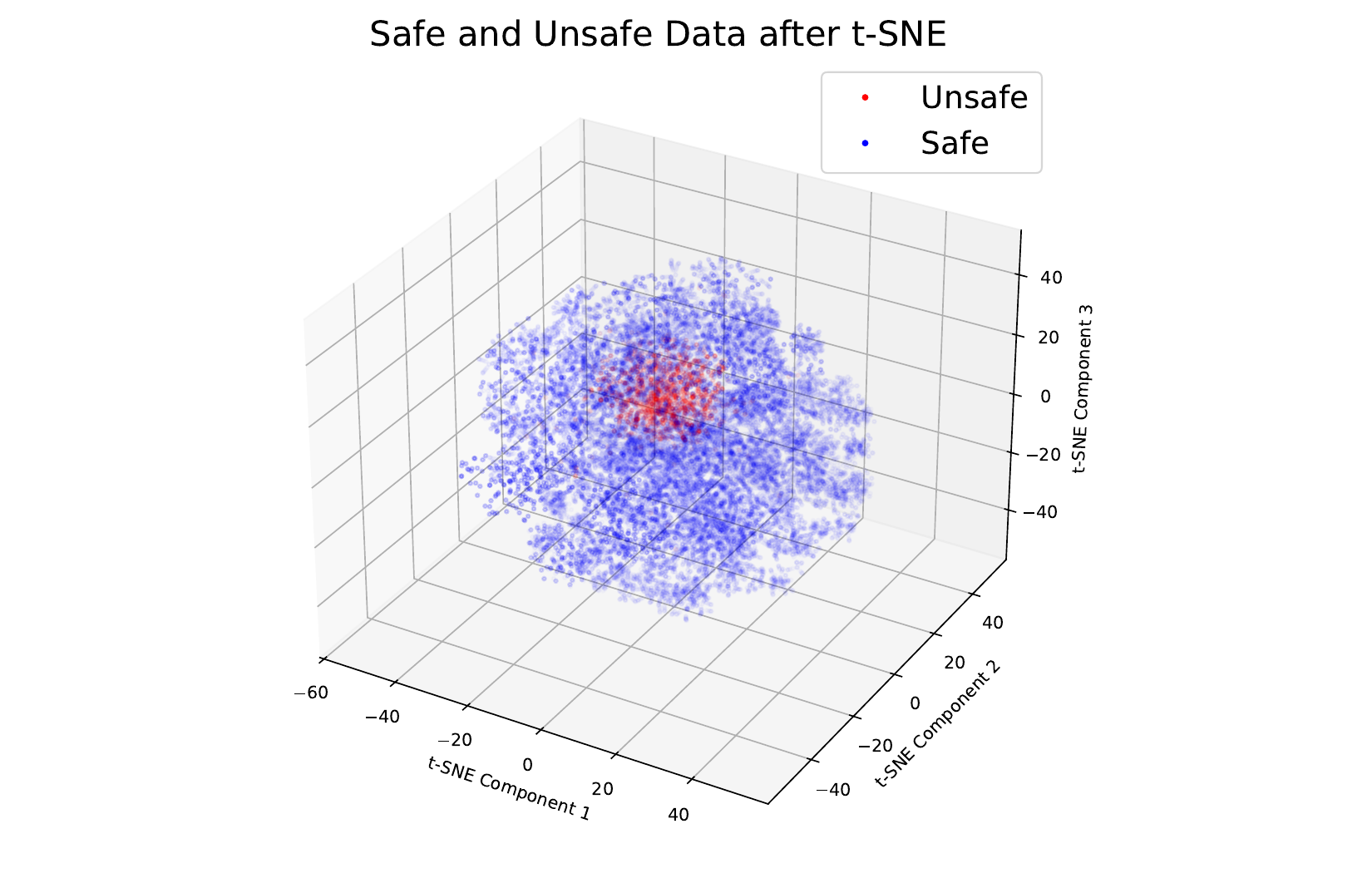}
    \caption{{\safediff} learns to differentiate between safe and unsafe phrases, with the two categories becoming clearly distinct after applying t-SNE for dimensionality reduction.}
    \label{distribution}
    \end{subfigure}
    \caption{Overview of {\safediff} (left). {\safediff} learns to distinguish safe and unsafe phrases (right).}
\end{figure*}

\paragraph{\textbf{Identifier Dataset}}
\label{L:Identifier dataset}
    
    As mentioned previously, we aim to detect inappropriate concepts in prompts to avoid inappropriate content generation. The first step in our pipeline is to construct the dataset with unsafe terms capturing the concepts of $C_{\text{unsafe}}$.
    To achieve this, we start by collecting open-sourced blacklisted phrases.
    Midjourney~\citep{midjourney} employs a blacklist of words and phrases that includes phrases associated with violence, hate speech, explicit sexual content, illegal activities, and other categories considered inappropriate by the platform. Additionally, Latent Guard~\citep{liu2024latent} introduces the CoPro dataset, which includes safe and unsafe prompts centered around blacklisted concepts. We build our blacklist based on Midjourney and CoPro dataset.
    
    Although we can collect numerous NSFW terms from open-source datasets, these datasets may be imbalanced and could lack some categories defined in $C_{\text{unsafe}}$.
    For instance, the amount of data in categories involving shocking or illegal activities is lower than in other categories.
    To tackle this, we leverage an LLM to generate related terms $t_c$ centered around one sampled concept $c$ in the blacklist of $C_{\text{unsafe}}$ as illustrated in \autoref{F:ar}, similarly to~\citet{hammoud2024synthclip} and \citet{liu2024latent}. This allows us to create a set $T_{\text{unsafe}} = \{\;t_{c}\;|\; c \in C_{\text{unsafe}}\}$.
    The unsafe terms in $T_{\text{unsafe}}$~(Red block on the left) simulate typical unsafe terms that a malicious user may input.
    In addition to the unsafe terms, we combine randomly selected 500 prompts (Blue block on the left) from the COCO 30k dataset~\citep{lin2014microsoft} and the following mentioned paired safe phrases dataset (Green block on the left) to serve as our safe prompt dataset. All phrases within these prompts are considered safe.

    \paragraph{\textbf{Steer Model Dataset}}
    \label{L:Steer dataset}
    In our subsequent steering transformation training procedure, we synthesize additional safe terms to steer unsafe embeddings toward safe ones. The core idea is to associate each unsafe term $t_c \in T_{\text{unsafe}}$  with a corresponding safe term ${t_c}'$ of similar meanings, allowing us to convert unsafe concepts into safe alternatives while preserving the original semantic intent of the prompt.
    For example, consider the term ``killed" in the prompt ``A man got killed.'' which represents a violent visual scene linked to the concept of ``violence''. As demonstrated in \autoref{F:ar}, we use GPT-4~\cite{openai2023chatgpt} to eliminate unsafe concepts in the input term $t_c$. In this case, a possible safe term ${t_c}'$ would be ``saved", transforming the meaning of the prompt to something like ``A man got saved”.
    By processing all elements in $T_{\text{unsafe}}$, we generate paired safe phrases dataset $T_{\text{safe}}$~(Green block on the left), comprising $M$ safe terms ${t_c}' \in T_{\text{safe}}$. Statistics of the dataset are listed in supplementary materials.

\vspace{-0.15cm}
\subsection{Inappropriate Concepts Identifier}
    \label{identifier}

    To ensure the generated images adhere to safety guidelines, we detect and mitigate undesirable concepts within user prompts before they are processed by the text-to-image diffusion model. The core idea is that the prompt embeddings can be leveraged to represent individual terms or phrases, enabling precise identification of inappropriate content.

    As shown in \autoref{F:ar}, both safe and unsafe phrases are first embedded using a text encoder and then utilized to train the identifier. The goal is for the identifier to accurately detect unsafe concepts as defined in \autoref{unsafeconcept}. Let \( T_{\text{unsafe}} \) denote the set of unsafe phrases. Specifically, the set of safe phrases consists of two subsets: \( T_{\text{safe}} \), containing synthetic phrases, and \( T_{\text{safe}'} \), comprising phrases collected from open-source resources. 
    
    The identifier's task is to classify a given phrase as either unsafe or safe. Let the input phrase embeddings be denoted as \( E_i \), the identifier's output as \( \hat{y}_i \), and the true label as \( y_i \in \{0, 1\} \), where \( y_i = 1 \) indicates an unsafe phrase and \( y_i = 0 \) denotes a safe phrase. The identifier is optimized using binary cross-entropy loss:
    \begin{equation}
        \mathcal{L} = -\frac{1}{N} \sum_{i=1}^{N} \left[ y_i \cdot\log \hat{y}_i + (1 - y_i) \cdot\log (1 - \hat{y}_i) \right]
    \end{equation}
    where $N$ is the total number of phrases in the training set.

    Since embeddings with similar semantics are known to cluster in the embedding space~\citep{mikolov2013distributed,radford2021learning}, we expect unsafe embeddings to naturally aggregate; thus, {\safediff} can demonstrate a generalization capability.
    As demonstrated in \autoref{distribution}, a t-SNE visualization shows a clear separation between safe and unsafe embedding spaces derived from MMA~\citep{yang2024mma} and COCO~\citep{lin2014microsoft} datasets.
    Moreover, the experimental results in \autoref{exp2} further confirm that {\safediff} maintains robust performance against various red-teaming attacks.

    In practice, we employ a lightweight multi-layer perceptron model to classify the embeddings of individual terms. Intuitively, we want to identify the phrase that precisely includes the inappropriate concept. For example, in the prompt ``a man got shot." only the term ``shot" relates to the concept of violence. Additionally, the phrase ``screw you" is related to the inappropriate concept ``hate", but neither ``screw" nor ``you" is related to the concept ``hate". To address such cases, we utilize a sliding identification technique to identify single terms and phrases that may collectively express inappropriate concepts.

\vspace{-0.15cm}
\begin{algorithm}
    \caption{Sliding Identification}
    \begin{algorithmic}[1]
    \State \textbf{Initialize:} window size $w$, position map $\mathcal{P}$, token length $L$, prompt vector set $\mathcal{V}$
    \For{$i = 1$ to $L$} 
        \For{$j = 1$ to $w$}
            \State $t \gets \min(L, i+j)$ \Comment{Set tail index}
            \State Extract token embeddings $e[i:t]$
            \State Pad to fixed size and store in $\mathcal{V}$
            \State Map token span in $\mathcal{P}$
        \EndFor
    \EndFor
    \State Predict unsafe probabilities for $\mathcal{V}$ using identifier model
    \end{algorithmic}
    \label{alg:sliding_window}
\end{algorithm}
\vspace{-0.15cm}

    \paragraph{\textbf{Sliding Identification}}
    To detect inappropriate concepts with fine-grained precision, we introduce a sliding identification mechanism that captures distributed unsafe signals across contiguous token spans. This method systematically scans a prompt using a sliding window over token embeddings, allowing the model to identify not only individual offensive terms but also compound phrases whose semantic meaning emerges only when considered jointly. As illustrated in \autoref{alg:sliding_window}, for each starting token in the prompt, we construct multiple token spans of varying lengths up to a predefined window size $w$. Each span is embedded, padded to a fixed size, and evaluated by the identifier to assess if it contains unsafe content.

\subsection{Steering Toward Safe Content}
\label{steer}
    To guide the generation process toward safe outputs, we propose a transformation that maps unsafe prompt embeddings to safe regions while preserving their original semantic meaning. Our key insight is that linear transformations in the word embedding space can effectively alter the generation style, as demonstrated in prior work~\citep{han2023lm}. Since {\safediff} operates directly on these embeddings, this approach allows us to adjust unsafe representations before they are processed by the diffusion model.

    Formally, let \( e_{\text{unsafe}} \) denote the embedding of an unsafe phrase. We define the steered embedding as:
    \begin{equation}
        e_{\text{steered}} = \epsilon \cdot W \cdot e_{\text{unsafe}} + (1 - \epsilon) \cdot e_{\text{unsafe}}
        \label{EQ: steer}
    \end{equation}
    where \( W \) is a learnable transformation matrix and \( \epsilon \in [0, 1] \) is a hyperparameter controlling the intensity of the transformation. This formulation allows for fine-grained control, interpolating between the original and transformed embeddings.
    
    To learn the transformation matrix \( W \), we utilize a supervised learning framework on a paired dataset consisting of unsafe phrases and their corresponding safe counterparts, as described in \autoref{traindatacollect}. The training objective minimizes the discrepancy between the transformed embedding and the target safe embedding:
    \begin{equation}
        \mathcal{L} = \left\| e_{\text{safe}} - W \cdot e_{\text{unsafe}} \right\|^2
    \end{equation}
    where \( e_{\text{safe}} \) is the embedding of the corresponding safe phrase in the paired dataset. This supervised learning framework ensures that the transformation learns meaningful semantic shifts toward safety, enabling the system to retain user intent while suppressing unsafe visual outputs.

    Although the transformation matrix \( W \) is trained on curated phrase pairs, the steering mechanism is designed to be modular and efficiently updatable. As safety policies or blacklists evolve, new safe–unsafe pairs can be incrementally added without requiring full retraining from scratch. Due to the relatively low dimensionality and simplicity of the linear transformation, \( W \) can be re-optimized rapidly using a small number of new examples (e.g., via fine-tuning or few-shot adaptation). This allows the system to dynamically accommodate new safety requirements with minimal engineering overhead, making it practical for deployment in fast-changing moderation contexts.

    While our framework employs a unified transformation matrix $W$ to steer unsafe prompt embeddings, it is important to clarify that $W$ encapsulates a multidimensional safety subspace that naturally attenuates various unsafe concepts.
    In our approach, the sliding window mechanism identifies unsafe segments within a prompt, and the same transformation is applied uniformly. As a result, while the framework is effective in steering the overall unsafe content toward safe regions, it does not currently support dynamic, selective suppression of individual unsafe aspects (e.g., reducing violent content while retaining nudity-related elements). We acknowledge this as a limitation of our fixed-$\epsilon$ design and consider the development of concept-specific steering strategies as future work.


\subsection{Inference}
\label{inference}
    Once {\safediff} is trained, it can be seamlessly integrated into diffusion models without additional fine-tuning. In practical applications, {\safediff} effectively detects the presence of blacklisted concepts and steers inappropriate prompts toward generating safe images.

    Consider a T2I model equipped with a text encoder. As illustrated in \autoref{F:overviewsafe}, a user provides a prompt, which can be either safe or unsafe. The input prompt is first embedded by a text encoder. 
    We define a concept blacklist $C_{\text{unsafe}}$ that contains potentially inappropriate concepts. 
    Then the identifier detects whether prompt $p$ contains any inappropriate concepts $c' \in C_{\text{unsafe}}$. If such a concept is detected, a linear transformation is applied to the prompt’s embedding to steer it toward safer content.
    Specifically, this transformation alters the latent representation to avoid generating inappropriate content without compromising the semantics of the original prompt. Once the transformation is applied, the modified embedding is passed through the diffusion model to generate a safe image. The transformed embedding ensures that the resulting image adheres to Safe-for-work guidelines while preserving the user’s original intent.

    Our method offers three key advantages: efficiency, effectiveness, and versatility. By operating at the prompt embedding level, it eliminates the need for computationally intensive model retraining while preserving the original semantics, allowing {\safediff} to act as an intermediary between user input and the diffusion model. This approach prevents unsafe content at the earliest stage of generation, blocking unsafe latent representations directly and reliably. Additionally, this steering process is lightweight and easily adaptable, enabling {\safediff} to align generated images with ethical and safety standards without introducing significant delays in the image generation pipeline. The transformation can be dynamically updated as blacklists or safety criteria evolve, making the system robust and scalable for various concept removal tasks in real-world applications.

\section{Experiment}
\label{sec:exp}


\subsection{Benchmarks and Metrics}
    We evaluate our safety mitigation technique using multiple benchmarks that assess both the reduction of inappropriate content and the maintenance of image quality and text alignment. Lastly, we analyze \safediff's versatility via an artist style removal task.
    
    \paragraph{\textbf{Inaproppriate Image Prompts~(I2P)}}
    The I2P benchmark~\cite{schramowski2023safe} consists of 4,703 real user prompts likely to produce inappropriate images. We measure performance using two metrics: \emph{Inappropriateness Rate:} The ratio of inappropriate images out of all images generated with the vanilla method. \emph{Attack Success Rate (ASR):} The percentage of images flagged as inappropriate when adversarial red-teaming techniques are applied.
    
    \paragraph{\textbf{MS-COCO FID-30K}}
    We adopt the MS-COCO FID-30K benchmark~\citep{boomb0omT2IBenchmark}, which comprises 30,000 captions sampled from the MS-COCO validation set~\citep{lin2014microsoft}. This benchmark has become a standard for evaluating text-to-image models. Performance is quantified using the Fréchet Inception Distance (FID) and CLIP scores, which respectively measure image fidelity and the semantic alignment between generated images and input prompts.
    
    \paragraph{\textbf{Artist Style Removal}}
    To demonstrate versatility, we evaluate our method on an artist style removal task following the setup from ESD~\citep{gandikota2023erasing}. The artist removal dataset comprises five modern artists and thematic styles: Kelly McKernan, Thomas Kinkade, Tyler Edlin, Kilian Eng, and the series ``Ajin: Demi-Human". For each artist, 40 generic prompts are composed to invoke the artist’s style. We assess style removal using the Learned Perceptual Image Patch Similarity (LPIPS) metric~\cite{zhang2018perceptual}. Specifically, LPIPS$_e$ is calculated on the erased artist. A higher LPIPS$_e$ value indicates more effective style removal. LPIPS$_u$ is calculated on unerased artists, with a lower LPIPS$_u$ suggesting minimal impact on unrelated artists.

\subsection{Models and Baseline}
    To evaluate the efficacy of our safety mitigation technique, we employ Stable Diffusion (SD) v1.4 as the base model. SD v1.4 is widely adopted in prior works on concept removal and red-teaming, making it the most relevant benchmark. For each prompt, three images are generated to reduce randomness, as generating only one image might coincidentally omit inappropriate content. We define a general unsafe concept set $C_{\text{unsafe}}$, applicable across all categories in I2P (details refer to supplementary materials). 

    We evaluated inappropriate image content by combining two classifiers. First, Q16 classifier~\citep{schramowski2022can}, which is also used for annotating LAION-5B dataset~\citep{schuhmann2022laion}, was utilized to detect a broad range of inappropriate content. Second, we applied NudeNet~\citep{nudenet} to identify sexually explicit content. For nude content, we classified exposed buttocks, exposed breasts, exposed genitalia, and exposed anus 
    content as inappropriate, while allowing otherwise provocative images~\citep{schramowski2023safe,wu2024erasediff,ma2024jailbreaking}. Unless otherwise specified, an image was classified as inappropriate if at least one of the classifiers flagged it as containing such content.
    
    \paragraph{\textbf{Compare Methods}}
    For the following \textbf{RQ1}, we evaluated {\safediff} with the state-of-the-art defense model SPM~\cite{lyu2024one}, Erased Stable Diffusion~(ESD)~\citep{gandikota2023erasing}, MACE~\cite{lu2024mace}, RACE~\cite{kim2024race}, AdvUnlearn~\cite{zhang2024defensive} and Safe Latent Diffusion~(SLD)~\cite{schramowski2023safe} under its MAX and STRONG configurations. Note that SPM, ESD, MACE, RACE, and AdvUnlearn can only defend the generation of nude images in their implementation.
    For \textbf{RQ2}, we selected state-of-the-art red-teaming frameworks, including Prompting4Debugging (P4D)\cite{chin2023prompting4debugging}, SneakyPrompt\cite{yang2024sneakyprompt}, Ring-A-Bell~\cite{tsai2023ring}, and MMA-Diffusion~\cite{yang2024mma}. Following the developers’ standard instructions, we conducted our evaluation on the I2P dataset with the P4D testbed, Ring-A-Bell nude, MMA-nudity, and NSFW\_200 dataset~\cite{yang2024sneakyprompt}, utilizing the standard configurations provided by each framework.

\subsection{RQ1: Effectiveness of {\safediff}}
    To investigate the ability of {\safediff} in both identifying and steering inappropriate concepts, we started by demonstrating its effectiveness in reducing the generation of explicit content.
    Next, we expanded the scope of inappropriate concepts to $C_{\text{unsafe}}$ to investigate whether {\safediff} could effectively identify and steer prompts containing a wider range of inappropriate content toward generating safe images. 
    Notably, we used inappropriate terms from {\safediff}'s training set as the naive blacklist.

\begin{table}[ht]
    \centering
    \caption{{\safediff} effectively removes sexual content from SD v1.4 on the I2P dataset, outperforming defend methods ESD~\cite{gandikota2023erasing}, SPM~\cite{lyu2024one}, MACE~\cite{lu2024mace}, AdvUnlearn~\cite{zhang2024defensive}, SLD STRONG, and SLD MAX~\cite{schramowski2023safe}.}
    \begin{tabular}{lc}
    \toprule
    \multirow{2.5}{*}{\textbf{Method}} & \textbf{Inappropriate probability \% $\downarrow$}\\
    \cmidrule{2-2}
    & I2P(Sexual)~\cite{schramowski2023safe}\\
    \midrule
    SD-v1.4 & 46.29 \\
    ESD~\cite{gandikota2023erasing} & 37.52\scriptsize(-8.77) \\
    SPM~\cite{lyu2024one}  & 23.39\scriptsize(-22.90) \\
    SLD STRONG~\cite{schramowski2023safe}  & 32.65\scriptsize(-13.64) \\
    SLD MAX~\cite{schramowski2023safe}  & 21.04\scriptsize(-25.25) \\
    MACE~\cite{lu2024mace} & 17.43\scriptsize(-28.86)\\
    RACE~\cite{kim2024race} & \textbf{10.22\scriptsize(-36.07)}\\
    AdvUnlearn~\cite{zhang2024defensive} & 21.13\scriptsize(-25.16)\\
    {\safediff}~(Ours) & \underline{16.33\scriptsize(-29.96)} \\
    \bottomrule
    \end{tabular}
    \label{T:nude}
\end{table}

\begin{table*}[]
    \caption{ {\safediff} demonstrates the best performance in reducing the probability of generating inappropriate content  (where lower values are better). The probabilities shown represent the likelihood of generating images classified as inappropriate by combining the Q16 and NudeNet classifiers across various I2P categories. The best performances are bolded, and the second-best performances are underscored.}
    \begin{minipage}{\linewidth}
    \centering
    \begin{tabular}{llllllll|c}
    \toprule
    \multirow{2.5}{*}{\textbf{Method}} & \multicolumn{8}{c}{\textbf{Inappropriate probability~\% $\downarrow$}}\\
    \cmidrule(lr){2-9}
    & \textbf{hate} & \textbf{harassment} & \textbf{violence} & \textbf{self-harm} & \textbf{sexual} & \textbf{shocking} & \textbf{illegal activity} & \textbf{Overall}\\
    \midrule
    SD-v1.4 & 27.27 & 19.05 & 27.65 & 30.34 & 46.29 & 35.98 & 18.16 & 30.17\\
    \midrule
    Blacklist & 19.48\scriptsize(-7.79) & 14.68\scriptsize(-4.37) & 17.99\scriptsize(-9.66) & 18.23\scriptsize(-12.11) & 21.59\scriptsize(-24.70) & 22.31\scriptsize(-13.67) & 10.87\scriptsize(-7.29) & 17.86\\

    SLD STRONG & 6.49\scriptsize(-20.78) & 6.80\scriptsize(-12.25) & 5.42\scriptsize(-22.23) & 5.24\scriptsize(-25.10) & 12.67\scriptsize(-33.62) & 11.33\scriptsize(-24.65) & \underline{3.03\scriptsize(-15.13)} & 7.97\\
    SLD MAX & \textbf{3.90\scriptsize(-23.37)} & \textbf{3.76\scriptsize(-15.29)} & \underline{5.29\scriptsize(-22.36)} & \textbf{2.25\scriptsize(-28.09)} & \underline{8.38\scriptsize(-37.91)} & \textbf{6.31\scriptsize(-29.67)} & \textbf{2.75\scriptsize(-15.41)} & \underline{5.15}\\
    \midrule
    {\safediff}~(Ours) & \underline{5.63\scriptsize(-21.64)} & \underline{4.25\scriptsize(-14.80)} & \textbf{2.91\scriptsize(-24.74)} & \underline{4.74\scriptsize(-25.60)} & \textbf{6.89\scriptsize(-39.40)} & \underline{6.78\scriptsize(-29.20)} & 3.99\scriptsize(-14.17) & \textbf{4.51}\\
    \bottomrule
    \end{tabular}
    \end{minipage}
    \label{T:exp1}
\end{table*}

    \paragraph{\textbf{Explicit Content Removal}}
    We first evaluate the performance of {\safediff} against competitive methods on the I2P dataset, focusing on the sexual content category. \autoref{T:nude} reports the inappropriate probability for each method, where the values in parentheses denote the percentage reduction with respect to the SD-v1.4 baseline. The baseline SD-v1.4 model exhibits an inappropriate probability of 46.29\%. Post-hoc methods, such as ESD and SPM, reduce this probability to 37.52\% and 23.39\%, respectively. More advanced techniques like SLD MAX, MACE, and RACE further lower the probability. RACE achieves the best result with 10.22\%. Meanwhile, our proposed {\safediff} obtains a value of 16.33\%, representing a significant reduction (29.96\%). These quantitative results demonstrate that {\safediff} effectively mitigates explicit content generation.
    
    \paragraph{\textbf{Inappropriate Content Removal}}
    We further investigated a more comprehensive inappropriate set $C_{\text{unsafe}}$ defined in \autoref{unsafeconcept}. We began our evaluation by demonstrating the inappropriate generation of SD v1.4 without any safety measures, as well as a basic blacklist-based prompt filtering approach. \autoref{T:exp1} presents the probability of generating inappropriate content for each category. 
    Varying from different categories, SD v1.4 generated inappropriate content with probabilities ranging from 18.16\% to 46.29\%. The naive blacklist approach slightly reduced probability, but inappropriate content was still generated in 17.86\% of cases across all categories.
    This confirms that simple blacklists, while providing minimal mitigation, remain ineffective as comprehensive safeguards due to their inability to address the semantic complexity and evolving nature of unsafe content.

    Next, we evaluated our proposed method {\textsc{SafeDiff}} against state-of-the-art safety mechanisms SLD STRONG and SLD MAX across various I2P concepts. As shown in \autoref{T:exp1}, {\textsc{SafeDiff}} demonstrated superior performance, reducing inappropriate content generation by over 95\%, with SLD MAX ranking second in effectiveness.
    In particular, {\safediff} outperformed its closest competitor, SLD MAX, in categories of violence, sexual content, and overall inappropriate content. As a result, only 5\% of images generated by {\safediff} were still classified as inappropriate. However, it is worth noting that the Q16 and Nudenet classifiers tend to flag images as inappropriate even when problematic content has been significantly reduced.
    In summary, {\safediff} effectively mitigates the generation of inappropriate content through targeted identification and modification of unsafe concepts within the text embedding space, offering a more robust solution compared to existing approaches. 
    


\subsection{RQ2: Effectivenss of {\safediff} under red-teaming attacks}
\label{exp2}
    \textsc{SafeDiff} achieves state-of-the-art performance on the I2P dataset; robustness against red-teaming attacks remains a critical criterion for effective safety defense. Red-teaming strategies vary significantly in nature. Black-box attacks typically paraphrase or substitute unsafe content with semantically similar alternatives to evade detection, whereas white-box attacks leverage model gradients or internal representations to bypass safety mechanisms. To comprehensively assess robustness, we first evaluate \textsc{SafeDiff} under the black-box red-teaming attack \textsc{P4D}~\cite{chin2023prompting4debugging}. Subsequently, we analyze its performance against three representative white-box attacks: \textsc{SneakyPrompt}~\cite{yang2024sneakyprompt}, \textsc{Ring-A-Bell}~\cite{tsai2023ring}, and \textsc{MMA-Diffusion}~\cite{yang2024mma}.

\begin{table}[ht]
    \centering
    \caption{Attack success rate (\% $\downarrow$) on I2P and P4D benchmarks.}
    \begin{tabular}{lccc}
    \toprule
    \multirow{2.5}{*}{\textbf{Method}} & \multicolumn{3}{c}{\textbf{Attack success rate (ASR)~\% $\downarrow$}}\\
    \cmidrule(lr){2-4}
    & I2P (all) & P4D (nude) & P4D (all) \\
    \midrule
    SD-v1.4 & 30.17 & 95.21 & 98.70 \\
    ESD & - & 55.40\scriptsize(-39.81) & - \\
    SLD STRONG & 7.97\scriptsize(-22.20) & 48.21\scriptsize(-47.00) & - \\
    SLD MAX & \underline{5.15\scriptsize(-25.02)} & \underline{37.25\scriptsize(-57.96)} & \underline{30.95\scriptsize(-67.75)} \\
    {\safediff}~(Ours) & \textbf{4.51\scriptsize(-25.66)}& \textbf{25.36\scriptsize(-69.85)} & \textbf{29.16\scriptsize(-69.54)} \\
    \bottomrule
    \end{tabular}
    \label{T:black}
\end{table}

    \paragraph{\textbf{White-box Attack}}
    \autoref{T:black}
    To investigate the effectiveness of concept removal approaches, we first focused on the ``nudity" category, as it is commonly recognized as explicitly harmful in the context of generative models. Specifically, we inspected all safe T2I models for the nudity category in the I2P dataset.
    As shown in \autoref{T:exp2}, ESD, SPM, and SLD exhibited poor performance when countering nudity-related attacks. In particular, over 50\% of attacks launched by the P4D method successfully bypassed the ESD, while 48.21\% and 37.25\% circumvented SLD STRONG and MAX defenses, respectively. In contrast, {\safediff} achieves a significantly lower attack success rate of 25.36\%, marking it as the most effective defense against red-teaming attacks targeting nudity.

    Next, we evaluated the robustness of SLD MAX and {\safediff} across all categories within the I2P dataset under the P4D attack, excluding ESD and SLD STRONG due to limited category support. Results in \autoref{T:exp2} revealed that {\safediff} marginally outperformed SLD MAX, maintaining an Attack Success Rate of around 30\%. This discrepancy may be attributable to the broader and more ambiguous scope of larger-scale unsafe concepts, which complicates effective defense, as also observed by \citep{ma2024jailbreaking}. Nonetheless, {\safediff} remains the most competitive defense across all categories. Detailed quantitative results for each unsafe concept and image results are presented in the supplementary materials.

\begin{table}[ht]
    \centering
    \caption{Attack success rate (\% $\downarrow$) under SneakyPrompt, Ring-A-Bell, and MMA-Diffusion.}
    \fontsize{8.7}{11}\selectfont
    \begin{tabular}{lccc}
    \toprule
    \multirow{2.5}{*}{\textbf{Method}} & \multicolumn{3}{c}{\textbf{Attack success rate (ASR)~\% $\downarrow$}}\\
    \cmidrule(lr){2-4}
    & SneakyPrompt & Ring-A-Bell & MMA-Diffusion \\
    \midrule
    SD-v1.4 & 69.35 & 83.10 & 95.70 \\
    ESD & 51.00\scriptsize(-18.35) & 52.80\scriptsize(-30.30) & 87.30\scriptsize(-8.40) \\
    SPM & 34.00\scriptsize(-35.35) & \underline{34.18\scriptsize(-48.92)} & \underline{73.30\scriptsize(-22.40)} \\
    SLD STRONG & 14.50\scriptsize(-54.85) & 62.00\scriptsize(-21.10) & 92.00\scriptsize(-3.70) \\
    SLD MAX & \underline{8.50\scriptsize(-60.85)} & 57.50\scriptsize(-25.60) & 83.70\scriptsize(-12.00) \\
    {\safediff}~(Ours) & \textbf{7.50\scriptsize(-61.85)} & \textbf{27.85\scriptsize(-55.25)} & \textbf{28.80\scriptsize(-66.90)} \\
    \bottomrule
    \end{tabular}
    \label{T:whiteattack}
\end{table}

    \paragraph{\textbf{Black-box Attack}}
    Lastly, we assessed all safe T2I models under SneakyPrompt, Ring-A-Bell, and MMA-Diffusion. As shown in \autoref{T:whiteattack}, {\safediff} achieving 7.5\%, 27.85\%, and 28.80\% ASR, respectively. In comparison, ESD, SPM, SLD STRONG, and SLD MAX exhibited higher ASRs across all attack types. These results underscore the effectiveness of {\safediff} in defending against sophisticated red-teaming attacks.
    
    In summary, {\safediff} consistently outperforms other defense models across various datasets and attack methods, making it the most reliable approach for mitigating undesirable content generation in T2I diffusion models.

\begin{table}[ht]
    \centering
    \caption{Image fidelity and text alignment on COCO 30K.}
    \begin{tabular}{lcc}
    \toprule
    \multirow{3.5}{*}{\textbf{Method}} & \multicolumn{2}{c}{\textbf{MS-COCO FID-30K}}\\
    \cmidrule(lr){2-3}
     & Image Fidelity & Text Alignment\\
     & \textbf{FID-30K $\downarrow$} & \textbf{CLIP $\uparrow$}\\
    \midrule
    SD-v1.4 & - & \underline{0.81} \\
    ESD & 15.97 & \textbf{0.82} \\
    SPM & \textbf{13.24} & 0.80 \\
    SLD STRONG & 18.28 & 0.77 \\
    SLD MAX & 18.76 & 0.75 \\
    {\safediff}~(Ours) & \underline{15.45} & 0.78 \\
    \bottomrule
    \end{tabular}
    \label{T:coco}
\end{table}

\subsection{RQ3: Generated Images Quality}
    We have demonstrated the effectiveness of {\safediff} in mitigating the generation of inappropriate content in T2I diffusion models. However, maintaining high image fidelity and ensuring strong alignment between generated images and input text prompts are equally important. Ideally, {\safediff} should have minimal or no impact on prompts that are already safe. To assess these aspects, we evaluated {\safediff}  with MS-COCO FID-30K score for image fidelity and CLIP score for measuring alignment between generated images and input text prompts. 

    As shown in \autoref{T:coco}, {\safediff} achieves a FID-30K score of 15.45—the second-lowest among all baselines
    , indicating better image fidelity. While CLIP score (0.78) is slightly lower than ESD and SPM, it remains competitive among other methods, demonstrating that {\safediff} has minimal impact on text-image alignment specificity.

\subsection{Artist Style Removal}
\label{sec:dis}
\label{L:conceptforg}

In this section, we explore the potential of {\safediff} to erase specific artist styles. For this experiment, we follow the artist removal dataset from ESD~\cite{gandikota2023erasing}, where {\safediff} was applied to remove references to ``Van Gogh" and ``Kelly McKernan", while maintaining other artist styles, such as ``Pablo Picasso", ``Thomas Kinkade", ``Tyler Edlin", and ``Kilian Eng".

\begin{figure}[ht]
        \centering
        \includegraphics[width=\linewidth]{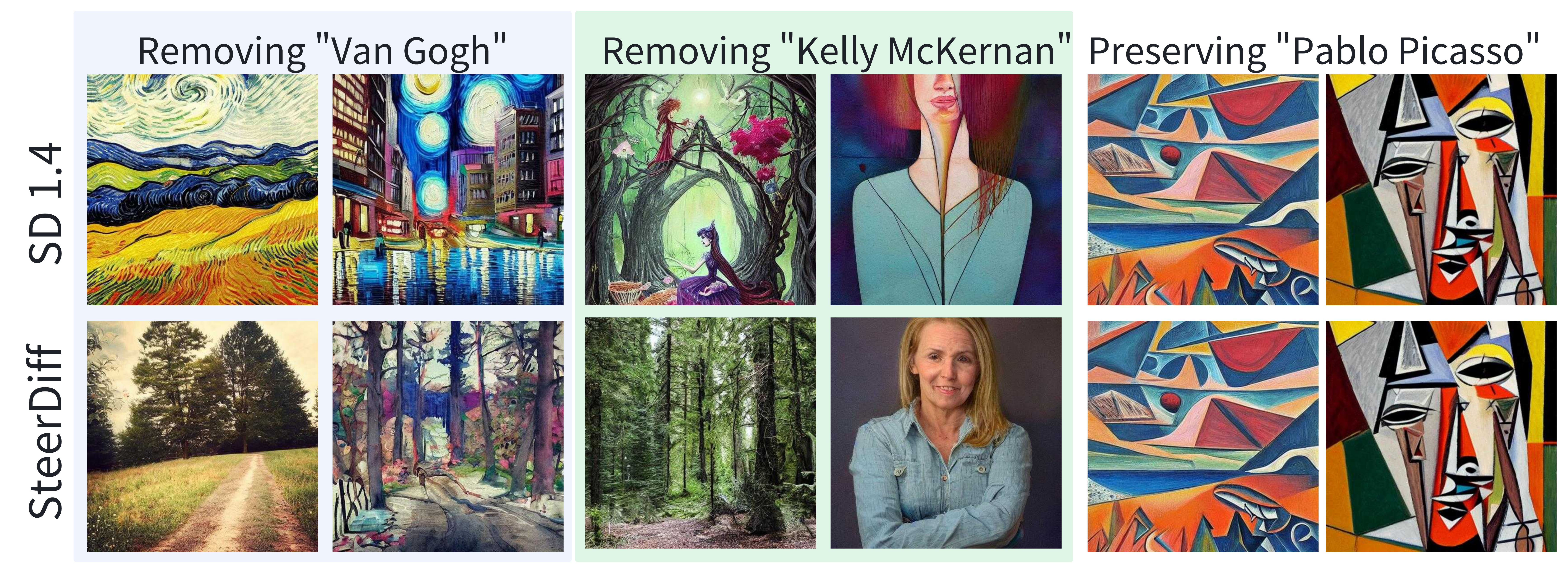}
        \caption{{\safediff} successfully removes the targeted concepts ``Van Gogh" and ``Kelly McKernan" while preserving unrelated concepts such as ``Pablo Picasso". The first row displays the original images generated by SD-v1.4, while the second row depicts steered samples generated from the same prompt.}
    \vspace{-0.5cm}
    \label{F:conceptforget}
\end{figure}

\begin{table}[ht]
    \centering
    \caption{Comparison of artist style removal. A higher LPIPS$_e$ value indicates more effective target style removal. A lower LPIPS$_u$ suggests minimal impact on unrelated artists.}
    \fontsize{8.6}{11}\selectfont
    \begin{tabular}{lcc|cc}
    \toprule
    \multirow{3}{*}{Method} & \multicolumn{2}{c|}{Remove “Van Gogh”} & \multicolumn{2}{c}{Remove “Kelly McKernan”} \\
    \cmidrule{2-5}
    & LPIPS$_e$ $\uparrow$ & LPIPS$_u$ $\downarrow$ & LPIPS$_e$ $\uparrow$ & LPIPS$_u$ $\downarrow$\\
    \midrule
        CA~\cite{kumari2023ablating} & 0.30 & 0.13 & 0.22 & 0.17 \\
        RECE~\cite{gong2024reliable} & 0.31 & 0.08 & 0.29 & 0.04 \\
        UCE~\cite{gandikota2024unified}& 0.25 & 0.05 & 0.25 & 0.03 \\
        {\safediff} (Ours) & \textbf{0.46} & \textbf{0.00} & \textbf{0.47} & \textbf{0.00}\\
    \bottomrule
    \label{T:dis}
    \vspace{-0.5cm}
    \end{tabular}
\end{table}

We follow the evaluation setup from RECE~\cite{gong2024reliable,zhang2018perceptual}, where LPIPS$_e$ is calculated on the erased artist. A higher LPIPS$_e$ value indicates more effective style removal. LPIPS$_u$ is calculated on unerased artists, with a lower LPIPS$_u$ suggesting minimal impact on unrelated artists. As shown in \autoref{T:dis}
and qualitative results in \autoref{F:conceptforget}, {\safediff} achieves superior performance. Our method not only effectively eliminates the target style but also preserves the integrity of non-target styles, by identifying steerable concepts and selectively modifying embeddings only when necessary.

\section{Related work}
\label{sec:relatedwork}

\subsection{Attacks on Text-to-Image Models}
    While T2I diffusion models have demonstrated impressive capabilities, they also pose potential risks related to the generation of harmful or inappropriate content. Red-teaming~\citep{ganguli2022red,perez2022red,li2024art,yang2024mma}, the practice of stress-testing AI systems to uncover vulnerabilities, has become a crucial area of research in ensuring the safety and ethical use of these models. In the context of T2I diffusion models, red-teaming~\citep{chin2023prompting4debugging,yang2024sneakyprompt,ma2024jailbreaking,zhang2023generate,zhuang2023pilot,tsai2023ring} focuses on detecting and mitigating the generation of undesirable or unsafe content, such as offensive imagery or representations of inappropriate concepts.

    Given that T2I diffusion models rely heavily on textual prompts to guide image generation, vulnerabilities often arise from the misalignment between the text embeddings and the visual content produced. Adversaries could potentially exploit these models by crafting malicious prompts that subtly bypass content filters. As a result, recent works have explored methods to steer these models away from generating unsafe outputs, such as by applying transformations to the prompt embeddings~\citep{han2023lm,li2025aligning,yoon2024safree,ahn2025distorting}, or by integrating ethical constraints into the generative process.
    The challenge lies in maintaining the model’s ability to generate diverse and creative images while ensuring that the output adheres to safety guidelines. Addressing this issue is essential for deploying these models in real-world applications, particularly in sensitive domains such as media, art, and content moderation.
    


\subsection{Towards Safe Image Generation}

    Current approaches to preventing undesirable images from generation generally follow three main strategies. The first involves removing undesired images from the training set, such as excluding all human figures~\citep{nichol2021glide} or selectively omitting specific undesirable image categories in the dataset~\citep{schuhmann2022laion,DALLE,stablediffusionv2}. However, this approach is costly as it necessitates retraining the model, and removing certain data classes often degrades the overall quality of the generated images~\citep{stablediffusion1v2}.
    The second strategy is post-hoc, includes using blacklists for blocking unsafe concepts~\citep{DALLE,midjourney,Leonardo,markov2023holistic}, 
    model editing methods~\citep{gandikota2023erasing,orgad2023editing,gandikota2024unified} and unlearning methods\citep{zhang2024forget,huang2024receler}. 
    Blacklists are straightforward to implement, they can be easily bypassed. Model editing methods identify and prune weights within the diffusion model to remove harmful concepts, but they often degrade output quality, thus harming the utility of the models~\citep{rombach2022high}. Unlearning methods fine-tune models to erase harmful content, but as model sizes grow, this becomes increasingly computationally expensive and less practical\citep{ramesh2022hierarchical,rombach2022high,saharia2022photorealistic}. Other approaches~\citep{park2024localization,zhang2024forget,yoon2024safree,gong2024reliable,zhang2024defensive, lyu2024one} use inpainting to mask unsafe content or attempt to unlearn inappropriate concepts either within the diffusion model~\citep{zheng2023imma} or the textual encoder~\citep{poppi2023removing}. These methods involve expensive fine-tuning, while {\safediff} offers an intermediary solution that operates between the input prompt and the diffusion model without additional training.

    A more promising direction involves training-free, filtering-based techniques. Latent Guard~\citep{liu2024latent} identifies inappropriate concepts prior to diffusion by learning a latent space on top of the T2I model’s text encoder; however, it does not actively remove these unsafe representations. In contrast, our proposed method, {\safediff}, acts as an intermediary between the input prompt and the diffusion model, steering unsafe concepts to safe regions without requiring additional model training. Our ablation studies demonstrate that the identifier in {\safediff} substantially outperforms Latent Guard: on the I2P dataset and MS-COCO FID-30K, our identifier achieves classification accuracies of 99.77\% and 99.90\%, compared to 30.13\% and 85.08\% for Latent Guard, respectively. Overall, while Latent Guard registers 58.43\% accuracy and 30.13\% recall, {\safediff} reaches 99.84\% accuracy and 99.77\% recall~(More details are listed in supplementary materials).
    Similarly, \citeauthor{wu2023uncovering} modifies text embeddings by replacing partial embeddings, but our method learns the transformation to steer unsafe concepts to safe while preserving semantics.
    Moreover, SteerDiff is explicitly designed for safety, incorporating sliding windows for fine-grained unsafe concept detection. This ensures its suitability for safety-driven t2i generation rather than attributes edition.

\vspace{-0.3cm}
\section{Discussion and Limitation}
In this work, we propose {\safediff}, a novel approach that identifies and steers inappropriate concepts in input prompts by intervening directly in the embedding space before the diffusion process begins. 
However, defining what constitutes ``inappropriate" is inherently subjective and varies across cultural and societal contexts, necessitating expert input and raising concerns about bias and fairness. These biases may be amplified by the training data, which often reflects the dominant social norms of specific groups. 

In addition to the ethical concerns discussed earlier, {\safediff} has certain technical limitations. {\safediff} currently uses a fixed steering intensity, which may result in overcorrection for borderline cases. A more dynamic and context-aware approach could mitigate this issue by adjusting the steering degree based on the severity of the unsafe content, ensuring more precise control without compromising the original semantics. Our evaluation focuses on a limited set of predefined concepts, suggesting the need for continual updates to remain aligned with evolving norms. 
Beyond content safety, {\safediff} shows potential for broader applications, such as artistic style removal, though ethical considerations must guide its deployment. In sum, while {\safediff} offers a promising paradigm for safer generative modeling, its effectiveness and fairness hinge on the careful handling of subjectivity, data biases, and ethical considerations.


\vspace{-0.3cm}
\section{Conclusion}
\label{sec:con}
The evolution of diffusion models in generating intricate images highlights both their potential and associated risks. Although recent research for diffusion models has made notable progress in mitigating inappropriate content generation, red-teaming studies reveal that these defenses can still be bypassed. Moreover, many defense strategies rely on fine-tuning the model to avoid generating inappropriate content, which becomes increasingly challenging as diffusion models grow larger.
In this paper, we present {\safediff}, a lightweight method that acts as an intermediary between the user’s input and the diffusion model, ensuring that generated images comply with ethical and safety standards. We conduct comprehensive experiments across various unlearning concepts to evaluate their effectiveness. Additionally, we benchmark {\safediff} using multiple red-teaming approaches to assess the robustness of our method. Lastly, we explore the potential of {\safediff} in artist-style removal tasks.


    
\bibliographystyle{ACM-Reference-Format}
\bibliography{main}

\clearpage
\setcounter{page}{1}

\section{Ethics Statement}
In this work, we introduce a paradigm to identify inappropriate concepts within input prompts and steer their embeddings to mitigate the generation of unsafe content. Unlike previous methods that primarily focus on post-hoc prevention or concept removal, {\safediff} operates directly on the embeddings of the input prompts, prior to the diffusion process. By intervening earlier in the generation pipeline, we aim to more effectively prevent the propagation of unsafe content.

However, the real-world application of {\safediff} requires carefully defining what constitutes inappropriate concepts, which may vary depending on the application domain. Defining these concepts is a non-trivial task and is likely to require input from human experts, potentially leading to subjective biases. These biases may stem from the social and cultural context in which the system is deployed, as the definition of inappropriateness is highly subjective and dependent on societal norms, which differ across regions and communities.
Moreover, since the notion of inappropriateness is largely defined by social norms, the system's performance may vary depending on whose norms are reflected in the training data. This introduces the risk of reinforcing the biases present in the data, as {\safediff} may disproportionately represent the values of the social groups most prominent in the training set.

Our testbed for evaluating inappropriateness is limited to specific, predefined concepts, which may not fully capture the diversity of opinions and sentiments regarding what is considered inappropriate. As societal norms evolve, so too must the definitions of inappropriateness used by {\safediff}. This necessitates regular updates to the training data and model parameters to ensure the continued relevance and fairness of the system.

Beyond the identification and mitigation of inappropriate content, we believe that {\safediff} can be applied to other areas, such as concept or artistic style removal. As discussed in \autoref{L:conceptforg}, {\safediff} has the potential to be extended to various applications, including the removal of specific artistic styles or other undesired concepts in generative models. However, such applications must also carefully consider the ethical implications of content modification, as indiscriminate use of these techniques could lead to censorship or the suppression of artistic expression.

In summary, while {\safediff} offers a promising approach to mitigating unsafe content generation, its reliance on subjective definitions of inappropriateness and the potential for reinforcing societal biases limit its scope and fairness.

\section{Reproducibility Statement}
Upon acceptance of this paper, all relevant code and data used in our experiments will be made publicly available. The repository will include the source code for {\safediff}, as well as the datasets and instructions necessary to reproduce the results. This will ensure transparency and encourage further research in this domain.

\section{Implementation details}
\subsection{Dataset Statistics}
\label{appendix: statistics of the dataset}
We incorporate 578 blacklist tokens sourced from the CoPro dataset. To augment these, additional unsafe expressions are generated through our LLM-based workflow, as illustrated in Figure 2a. The steer model is trained on 2,378 safe–unsafe pairs, while the identifier is trained using 19,012 safe and 2,378 unsafe phrases. Aligned with our goal of isolating and controlling specific unsafe concepts, we use only the blacklist tokens from CoPro, rather than the full prompts, when constructing training pairs.



\section{Training data generation}
    \label{datagen}
    Following the description in the main text, we start from defining a concept set $C_{\text{unsafe}}$ and generate related terms $t_c$ entered around one sampled concept $c$ in the blacklist of $C_{\text{unsafe}}$. As introduced in the paper, we do this by prompting an LLM for prompt generation. We follow Latent Guard in conditioning the T2I prompt. For a concept $c$, we prompt an LLM with the following text to generate $t_c$:
    
    \begin{tcolorbox}
        \textbf{Prompt for Generating $T_{\text{unsafe}}$:}
        You are a text-to-image diffusion model safety assistant. Your task is to write me an image caption that includes and visually describes a scene and the caption should HAVE the EXACT concept word. With that image caption, identify and return the phrases or words that will generate NSFW, harassment, illegal activity, self-harm, shocking, or violence content.
        Only return the phrase no other words, no other response.
    \end{tcolorbox}
    
    We also report the LLM prompt used to generate the paired safe phrases $T_{\text{safe}}$ corresponding to above unsafe phrases $T_{\text{unsafe}}$. The prompt is:
    \begin{tcolorbox}
        \textbf{Prompt for Generating $T_{\text{safe}}$:}
        You are a NSFW and SFW safety assistant. Return the safe and benign opposite meaning phrase of the given phrase. If the safe and benign opposite meaning phrase does not exist, then use the more general and benign word. Only return the phrase no other words, no other response.
    \end{tcolorbox}

\begin{figure}[ht]
        \centering
        \includegraphics[width=\linewidth]{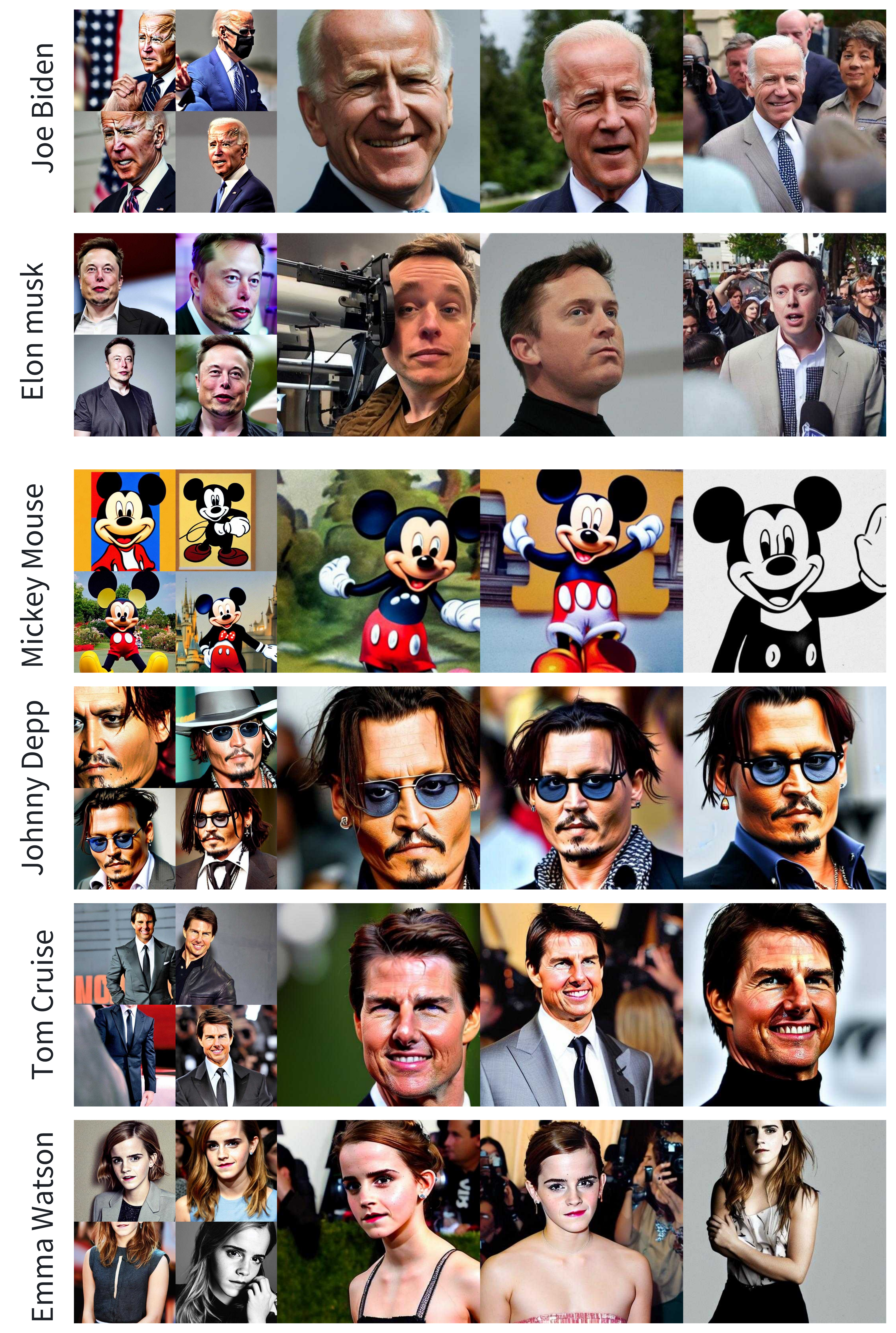}
        \caption{{\safediff} successfully removes the targeted concepts ``Van Gogh" and ``Kelly McKernan" while preserving unrelated concepts such as ``Pablo Picasso". The first row displays the original images generated by SD-v1.4, while the second row depicts steered samples generated from the same prompt.}
    \label{F:conceptforget}
\end{figure}
\section{Concepts removal}
     In this section, we explore the potential of {\safediff} to erase specific concepts during image generation. For this experiment, {\safediff} was applied to remove references to ``Elon Musk" and ``Joe Biden" from user inputs. As shown in \autoref{F:conceptforget}, the first row illustrates that {\safediff} successfully removed these concepts from the generated images. Notably, {\safediff} effectively removed target concepts while preserving some attributes of the original concepts, such as hairstyle and distinctive clothing style.
    We also evaluated the method’s effect on unrelated concepts, including ``Johnny Depp", ``Emma Watson", ``Tom Cruise", and ``Mickey Mouse". Ideally, {\safediff} should have little to no impact on unrelated concepts. As outlined in the last two rows, {\safediff} preserved these unrelated concepts. This demonstrates that our approach can be effectively applied to selective concept forgetting without affecting other content.

\section{Red-teaming attack analysis}
\label{L:redteaming}
    \begin{figure}[ht]
        \centering
        \includegraphics[width=\linewidth]{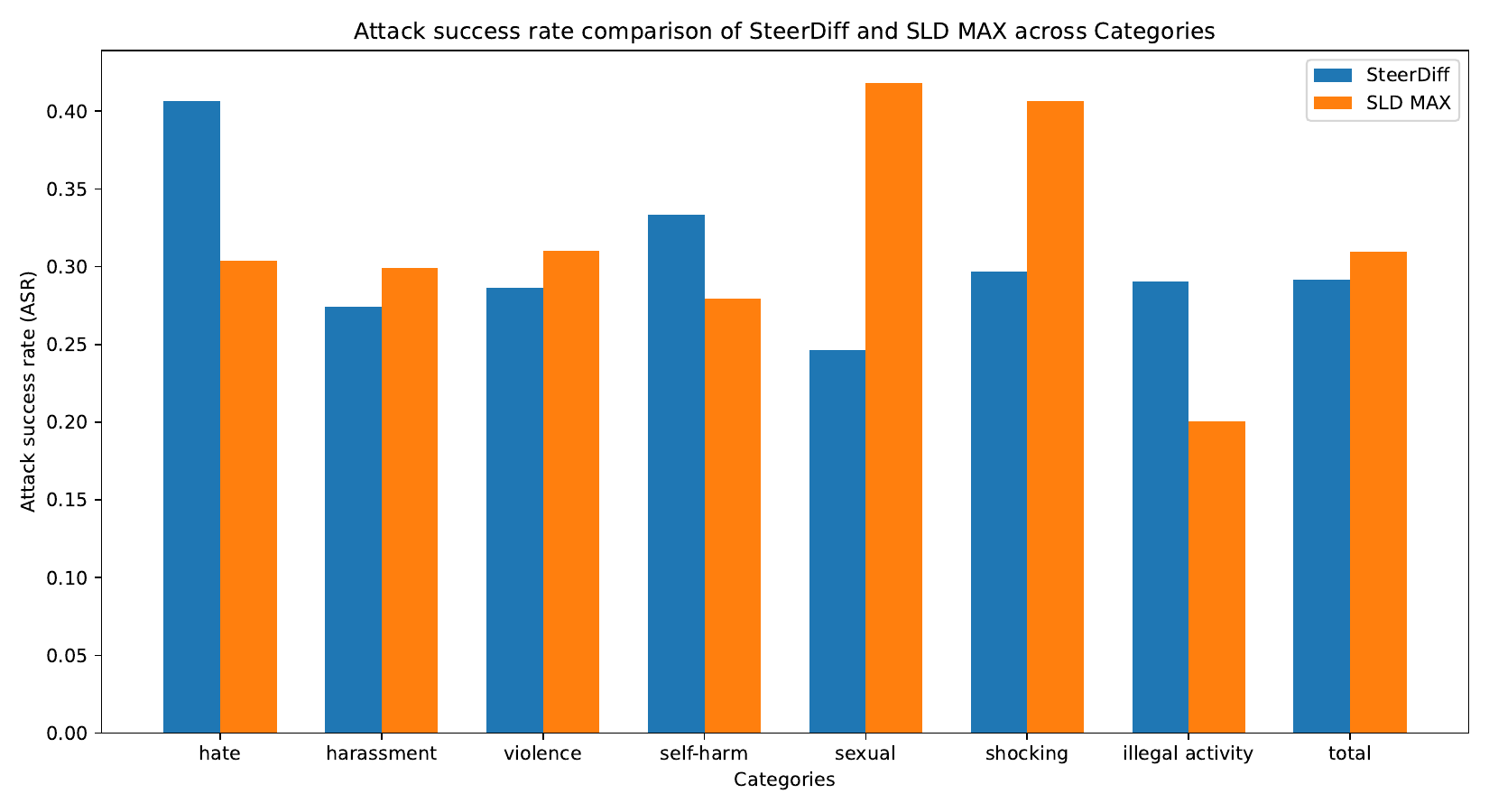}
    \caption{Attack success rate comparison of {\safediff} and SLD MAX across different categories (lower the better defense performance).}
    \label{F:p4d}
    \end{figure}
    
    \autoref{F:p4d} demonstrate the attack success rates (ASR) of two defense mechanisms, {\safediff} and SLD Max, across several categories on the I2P dataset. {\safediff} generally outperforms SLD Max in categories such as harassment, violence, sexual, and shocking, achieving lower ASR values in hate. In contrast, SLD Max demonstrates better defense ability in categories such as hate, self-harm, and illegal activity. Overall, the total ASR suggests a slight advantage for {\safediff} in terms of overall attack vulnerability.

\section{Visualization of steering}
\begin{figure}[ht]
        \centering
        \includegraphics[width=\linewidth]{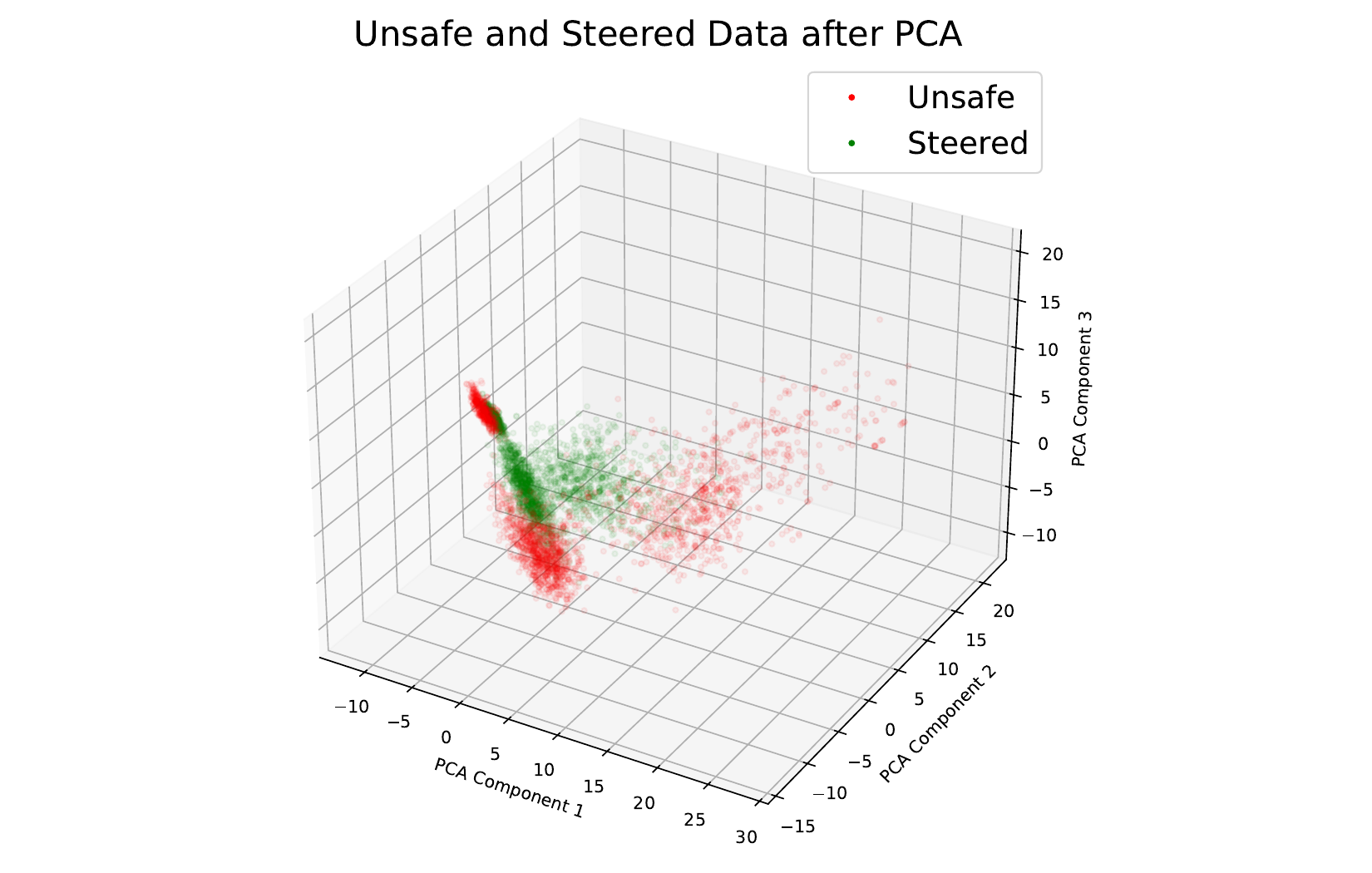}
    \caption{{\safediff} learns to project unsafe prompts.}
    \label{F:pca}
    \end{figure}

\autoref{F:pca} illustrates the outcome of applying the {\safediff}, which learns to project unsafe phrases into a safer latent space. In this 3D PCA visualization, the red points represent the original unsafe phrases, while the green points correspond to the steered versions of these phrases after transformation by SteerDiff.

\subsection{Examples of Generated Images}
In our evaluation of different models—SD v1.4, {\safediff}, ESD, SLD STRONG, and SLD MAX. We observe clear trade-offs between the ability to mitigate inappropriate content and the quality of generated images.

{\safediff} consistently demonstrates superior mitigation of inappropriate content while preserving the generative capabilities of the diffusion model. The images produced by {\safediff} maintain higher fidelity, with details and clarity closely resembling those from the baseline SD v1.4, making it the most balanced approach in terms of both safety and image quality.
On the other hand, ESD and SLD STRONG offer competitive image quality but occasionally fail to fully suppress inappropriate content generation. These models generate visually appealing images with relatively high quality, especially in complex textures and objects, but their inconsistency in filtering undesirable elements presents a notable shortcoming in safety-critical applications.

SLD MAX achieves a mitigation performance comparable to {\safediff} in terms of defense against inappropriate content generation. However, this comes at the cost of image quality, particularly in depictions of human faces. Images generated with SLD MAX tend to blur facial features, reducing the overall aesthetic quality. This blurring effect is less pronounced in non-human objects but remains a significant limitation in scenarios requiring fine detail preservation.

Overall, {\safediff} stands out as the most effective model for generating high-quality, safe content, whereas SLD MAX provides strong mitigation at the expense of visual detail, particularly in more nuanced areas like facial generation.

    \subsection{Examples of Generated Images in COCO-30K}
    \autoref{F:coco} illustrate the generated example of COCO-30k by Stable Diffusion v1.4, {\safediff}, ESD, SPM, SLD STRONG and SLD MAX.
    \begin{figure*}[ht]
        \centering
        \includegraphics[width=\textwidth]{fig/coco.pdf}
    \caption{Examples of images generated using COCO-30k prompts. From left to right, the columns represent the augmented prompts, images generated by {\safediff}, ESD, SPM, SLD STRONG, and SLD MAX, respectively.}
    \label{F:coco}
    \end{figure*}

    \subsection{Examples of Generated Images in I2P}
    \autoref{F:i2p} illustrate the generated example of I2P by Stable Diffusion v1.4, {\safediff}, ESD, SPM, SLD STRONG and SLD MAX.
    
    \begin{figure*}[ht]
        \centering
        \includegraphics[width=\textwidth]{fig/i2p-example.pdf}
    \caption{Examples of images generated using I2P prompts. From left to right, the columns represent the augmented prompts, images generated by {\safediff}, ESD, SPM, SLD STRONG, and SLD MAX, respectively. Red blocks have been added to the images to obscure explicit inappropriate content.}
    \label{F:i2p}
    \end{figure*}
    
    \subsection{Examples of Generated Images Under Sneakyprompt}
    \autoref{F:sneak} illustrate the generated example of I2P by {\safediff}, ESD, SPM, SLD STRONG and SLD MAX.
    \begin{figure*}[ht]
        \centering
        \includegraphics[width=0.9\textwidth]{fig/sneakyprompt-example.pdf}
    \caption{Examples of images generated using Sneakyprompt augmented prompts. From left to right, the columns represent the augmented prompts, and images generated by {\safediff}, ESD, SPM, SLD STRONG, and SLD MAX, respectively. Red blocks have been added to the images to obscure explicit inappropriate content.}
    \label{F:sneak}
    \end{figure*}
\label{L:appendiximage}

\newpage
\newpage

\end{document}